\documentclass{endm}

\usepackage{endmmacro}
\usepackage{graphicx}
\usepackage{epsfig}
\usepackage{amssymb}
\usepackage{algorithm}
\usepackage{algpseudocode}
\usepackage{array}
\usepackage{longtable}
\usepackage[table,xcdraw]{xcolor}
\usepackage{booktabs}
\usepackage{fancyhdr}
\usepackage{hyperref}
\fancypagestyle{firstpage}{%
	\fancyhf{} % sets both header and footer to nothing
	
	\lfoot{\scriptsize{\textcolor{blue}{{\fontfamily{lmr}\selectfont
					\textit{Please cite this paper as: }
					Antoniadis N. and Sifaleras A., "A hybrid CPU-GPU parallelization scheme of variable neighborhood search for inventory
					optimization problems", Electronic Notes in Discrete Mathematics, Elsevier B.V., Vol. 58, pp. 47-54, 2017.\\
					The formal publication is available at Elsevier via \url{http://dx.doi.org/10.1016/j.endm.2017.03.007}\\
				\textcopyright2017 Under CC-BY-NC-ND 4.0 license \url{http://creativecommons.org/licenses/by-nc-nd/4.0/} \\
	}
}}}
\cfoot{}
}

\begin{document}

% DO NOT REMOVE: Creates space for Elsevier logo, ScienceDirect logo
% and ENDM logo
\begin{verbatim}\end{verbatim}\vspace{2.5cm}

\begin{frontmatter}

\title{A hybrid CPU-GPU parallelization scheme of variable neighborhood search for inventory optimization problems}

\author[AngeloAddress]{Nikolaos Antoniadis\thanksref{Nikos_email} }
\author[AngeloAddress]{Angelo Sifaleras\thanksref{Angelo_email} }

\address[AngeloAddress]{Department of Applied Informatics, School of Information Sciences, \\
University of Macedonia, 156 Egnatia Str., Thessaloniki 54636, Greece}

\thanks[Nikos_email]{Email: \href{mailto:mai1521@uom.edu.gr} {\texttt{\normalshape mai1521@uom.edu.gr}}}
\thanks[Angelo_email]{Email: \href{mailto:sifalera@uom.gr} {\texttt{\normalshape sifalera@uom.gr}}}

\begin{abstract}
In this paper, we study various parallelization schemes for the Variable Neighborhood Search (VNS) meta–heuristic on a CPU-GPU system via OpenMP and OpenACC. A hybrid parallel VNS method is applied to recent benchmark problem instances for the multi-product dynamic lot sizing problem with product returns and recovery, that appears in reverse logistics and is known to be NP-hard. We  report our findings regarding these parallelization approaches and present promising computational results.
\end{abstract}


\end{frontmatter}
\thispagestyle{firstpage}
\section{Introduction}
\label{introduction}
Variable Neighborhood Search \cite{HMTH2016} is a simple, although efficient, metaheuristic method for the solution of various types of optimization problems. A systematic change of neighborhoods consists the core function of VNS, in order to intensify the search for better solutions and also to diversify local optimum solutions. VNS obtains a neighbor solution from the current solution and runs a local search procedure to reach a local optimum. If an improved solution is obtained through any of the neighborhood structures, then it is the new current solution. Otherwise, the perturbation (shaking) procedure searches for another neighboring solution to perform a new local search. 

All the state-of-the-art optimization solvers are already parallelized for exploiting the modern multiprocessor hardware systems. Similarly, researchers are motivated to parallelize their heuristic methods for performance tuning. Thus, one can find either parallel VNS heuristic implementations using either CPU parallelization schemes (MPI VNS \cite{DC2012}, OpenMP / threads VNS \cite{DPPS2013,SSRMD2015}), or even hybrid CPU-GPU VNS parallelization schemes \cite{CMOSBF2016}.

This paper considers a hybrid CPU-GPU parallelization of a variable neighborhood search method that was recently presented in \cite{SK2015,SK2016} for the solution of the uncapacitated multi-item economic lot-sizing problem with remanufacturing. 
%To the best of our knowledge there are only a few papers in the literature studying hybrid CPU-GPU VNS parallelization schemes.
The remainder of this note is organized as follows. Section \ref{problem} briefly describes the uncapacitated multi-item economic lot sizing problem with product returns and recovery. In Section \ref{methodology}, we present the research methodology and the limitations of our study. The experimental results are shown in Section \ref{results}. Finally, conclusions and future work are discussed in Section \ref{conclusions}.

%-------------------------------------------------------------------------------------------
%-------------------------------------------------------------------------------------------

\section{Inventory optimization problem}
\label{problem}

The uncapacitated multi-item economic lot-sizing problem with remanufacturing aims to compute i) the number of new $x_{M}(k,t)$ and/or remanufactured products $x_{R}(k,t)$ and also ii) the inventory level of serviceable items $y_{M}(k,t)$ and/or items that can be remanufactured $y_{R}(k,t)$, per period. The objective of the problem is to minimize the total cost due to i) manufacturing and/or remanufacturing setup cost, and also due to ii) holding cost for the serviceable items and/or recoverable items per unit time. The number of periods ($T$), different products ($K$), setup/holding costs ($k_{M}(k)$, $k_{R}(k)$, $h_{M}(k)$, $h_{R}(k)$), the demand for each time period and product ($D(k,t)$) and also the number of returned items per period and product ($R(k,t)$) that can be completely remanufactured and sold as new, constitute the parameters of this problem. The mathematical formulation of this problem is analytically presented in \cite{SK2016}.

%-------------------------------------------------------------------------------------------
%-------------------------------------------------------------------------------------------

\section{Research methodology and limitations} % Parallelization
\label{methodology}

The initialization method, and also the neighborhood structures of the proposed variable neighborhood search method are the same as those presented in \cite{SK2016} and are, thus, omitted in this short paper. Due to the construction of the neighborhood structures \cite{SK2016}, a strong data dependency was noticed by using the NVIDIA Visual Profiler. Thus, it was not possible to parallelize the ``find the best neighbor'' function (see Algorithm \ref{OpenACC_VND}) of the VND method. Therefore, the products-period loop was rather parallelized so that for each product we can, in parallel, calculate the new objective value (the problem is uncapacitated, and thus no link exist between the products). The only drawback is that the ``find the best neighbor'' function requires to be serially executed.

%\begin{algorithm}
%\caption{Parallel VND using OpenMP}
 %\label{OpenMP_VND}
%\footnotesize 
%\begin{algorithmic}
%\Procedure{VNDOpenMP}{$K$, $T$, $R$, $D$, $h_R$, $h_M$, $k_R$, $k_M$, $p_R$, $p_M$, $s$, $k_{max}$}
%\Repeat 
	%\State $improvement \gets 0$ 
	%\For{$k \gets 1, k_{max}$}
		%\State $!\$omp\ parallel\ private\ (s)$ \Comment{s: Value of the result}
		%\State $!\$omp\ do\ schedule(static)\ reduction(+:diff)$ \Comment{diff: Reduction variable}
		%\For{$i \gets 1, K$}
			%\For{$t \gets 1, T$}
		 		%\State $!\$omp\ critical$
		 		%\State Find best neighbor $s'_i$ of $s (s'_i \in N_k(s))$
		 		%\State $!\$omp\ end\ critical$		 		
		 	%\EndFor
		 	%\State $diff \gets diff+s-s'_i$
		%\EndFor
		%\State $!\$omp\ end\ do$
		%\State $!\$omp\ end\ parallel$
		%\State $s \gets s-diff$
		%\If{$diff > 0$}
			%\State $improvement \gets 1$
		%\EndIf
	%\EndFor
%\Until{$improvement = 0$}
%\EndProcedure
%\end{algorithmic}
%\end{algorithm}

\begin{algorithm}
\caption{Parallel VND using OpenACC}
 \label{OpenACC_VND}
\footnotesize 
\begin{algorithmic}
\Procedure{VNDOpenACC}{$K$, $T$, $R$, $D$, $h_R$, $h_M$, $k_R$, $k_M$, $p_R$, $p_M$, $s$, $k_{max}$}
\State $!\$acc\ data\ copy(<global\ variables>)$
\Repeat 
	\State $improvement \gets 0$ 
	\For{$k \gets 1, k_{max}$}
		\State $!\$acc\ parallel\ loop\ private(<private\ variables>)\ reduction(+:diff)\ gang\ vector$
		\For{$i \gets 1, K$}
			\State $!\$acc\ loop$
			\For{$t \gets 1, T$}
		 		\State Find best neighbor $s'_i$ of $s (s'_i \in N_k(s))$
		 	\EndFor
		 	\State $diff \gets diff+s-s'_i$
		\EndFor
		\State $s \gets s-diff$
		\If{$diff > 0$}
			\State $improvement \gets 1$
		\EndIf
	\EndFor
\Until{$improvement = 0$}
\State $!\$acc\ end\ data$
\EndProcedure
\end{algorithmic}
\end{algorithm}

Furthermore, we have tried OpenACC bearing in mind that our program is not compute-intensive and it has many memory transfers. However, using the $\$acc\ data\ copy$ directive several useless memory transfers were avoided and thus, this version managed to perform better than the four-threaded version OpenMP. Also, the parallelization strategy was exactly the same in both versions. 

\begin{figure}[H]
	\centering
		\includegraphics[width=\textwidth]{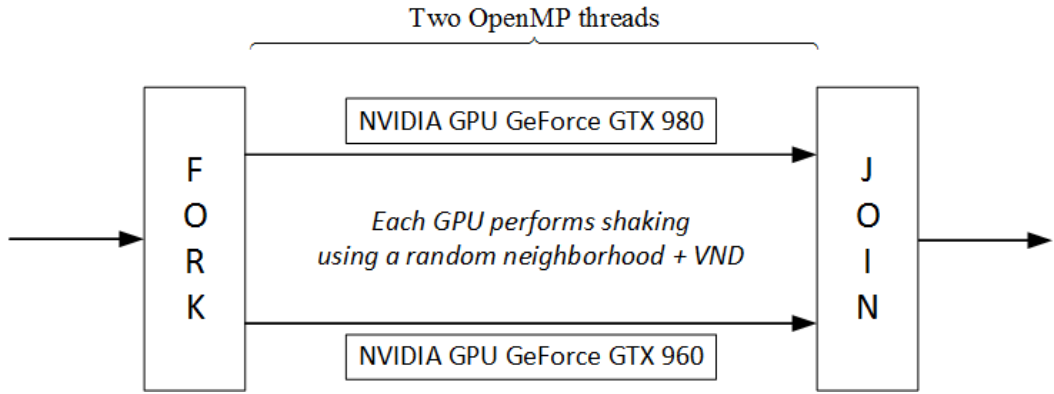}
	\caption{Fork-join model}
	\label{fork_join}
\end{figure}

Finally, the hybrid CPU-GPU version aims to execute in parallel the shaking and VND functions of the General Variable Search (GVNS) algorithm using OpenMP, on two GPUs (see Algorithm \ref{GVNS}). Each thread, which corresponds to a GPU, sends data from host memory to GPU and receives data from GPU to host memory (see Figure \ref{fork_join}). Afterward, the objective values of each thread are compared with each other, the minimum one is selected and the process continues until the time limit is reached. The VND function is using OpenACC as previously explained.

\begin{algorithm}
\caption{GVNS using two GPUs}
\label{GVNS}
\footnotesize 
\begin{algorithmic}
\Function{HybridGVNS}{$x$, $k'_{max}$, $k_{max}$, $t_{max}$}
\Repeat
	\State $k \gets 1$
	\Repeat
    	\State $!\$omp\ parallel\ private(tid)$
        	\State\hspace{\algorithmicindent} $tid \gets omp\_get\_thread\_num()$
            \State\hspace{\algorithmicindent} $call\ acc\_set\_device\_num(tid,\ acc\_device\_nvidia)$ 
            \State\hspace{\algorithmicindent} $!\$omp\ critical$
        	\State\hspace{\algorithmicindent}\hspace{\algorithmicindent} $x' \gets Shake(x, k)$
			\State\hspace{\algorithmicindent}\hspace{\algorithmicindent} $x'' \gets VND(x', k'_{max})$ 
			%\State\hspace{\algorithmicindent}\hspace{\algorithmicindent} $NeighbourhoodChange(x, x'', k)$
            %\State NOMIZW PWS DE XREIAZETAI H NeighbourhoodChange
            %\State DEN ALLAZOYME GEITONIA STH VNS. STH VND ALLAZOYME
            \State\hspace{\algorithmicindent} $!\$omp\ end\ critical$
        \State $!\$omp\ end\ parallel$
	\Until{$k = k_{max}$}
	\State $t \gets CpuTime()$
\Until{$t > t_{max}$}
\EndFunction
\end{algorithmic}
\end{algorithm}

Using PGI University Developer License the maximum available OpenMP threads were limited to four. Thus, it is possible that the OpenMP performance with eight OpenMP threads -equal to the processor's cores- could be improved than both OpenACC and hybrid solutions. 

%-------------------------------------------------------------------------------------------
%-------------------------------------------------------------------------------------------

\section{Experimental results}
\label{results}

The computational experiments were performed on a computer running Ubuntu Linux 14.04 64bit with an Intel Core i7 CPU 920 at 2.793 GHz with 8 MB SmartCache and 6 GB DDR3 400 MHz main memory. Also, the following two graphic cards were used: an NVIDIA GPU GeForce GTX 980 and an NVIDIA GPU GeForce GTX 960, on PCI Express x16 Gen2 bus. These GPUs featured 4 GB of memory and 2048 Cuda cores, and 2 GB of memory and 1024 Cuda cores, respectively. The VND implementation was implemented in Fortran and compiled using the PGI Accelerator Fortran/C/C++ Workstation for Linux - University Developer License.

The experimental computational study compared the performances of i) the serial implementation of VND presented in \cite{SK2016}, ii) an OpenMP implementation of GVNS, iii) an OpenACC implementation of GVNS, and finally iv) a hybrid OpenMP - OpenACC implementation of GVNS. The results presented in \cite{SK2016} were based on a VND and due to its serial implementation no shaking could be succesfully completed in a short amount of time. However, the new parallel implementations provided us the ability to successfully complete the shaking phase in the given time period and thus we now present more complete VNS schemes.

Each implementation was tested using a recent set of large benchmark instances presented in \cite{SK2016} with 300 products and 52 periods, and executed for 60 secs. The results of the comparative computational study are depicted in Table \ref{numerical_results}, where the best values per instance are denoted with bold font.

{\tiny %footnotesize
% Table generated by Excel2LaTeX from sheet 'Normal instances Parallel VNS'
\begin{longtable}{ccccc}
\label{numerical_results} \\
\hline
ID    & Serial VND & OpenMP GVNS & OpenACC GVNS & Hybrid OpenMP - OpenACC GVNS \\
\hline
\endhead
\hline
\endfoot
1     & 3270282.60 & 3265686.40 & 3265686.40 & \textbf{3264458.40} \\
2     & 3495448.50 & 3492634.00 & 3492014.00 & \textbf{3491153.50} \\
3     & 3748498.80 & 3746685.40 & 3745647.80 & \textbf{3745532.00} \\
4     & 4027173.00 & 4024427.20 & 4023575.00 & \textbf{4023025.0}0 \\
5     & 4479928.00 & 4479215.00 & \textbf{4479012.50} & 4479015.00 \\
6     & 4875797.60 & 4875022.40 & 4874594.40 & \textbf{4874358.80} \\
7     & 5796958.80 & 5793979.20 & \textbf{5793966.20} & 5794314.60 \\
8     & 6984412.00 & 6984088.00 & \textbf{6983409.50} & 6984130.00 \\
9     & 8081856.80 & 8080948.00 & 8080764.00 & \textbf{8080570.40} \\
10    & 4858495.80 & 4854200.60 & 4853070.00 & \textbf{4852411.60} \\
11    & 5089078.50 & 5085783.00 & \textbf{5084839.00} & 5085094.50 \\
12    & 5278825.40 & 5276336.80 & \textbf{5275247.00} & 5275955.80 \\
13    & 5892350.80 & 5884471.60 & \textbf{5882872.20} & 5883021.60 \\
14    & 6321038.50 & 6316290.00 & \textbf{6315141.00} & 6315600.00 \\
15    & 6704859.60 & 6701762.60 & \textbf{6700878.60} & 6702075.60 \\
16    & 7414346.80 & 7413521.60 & 7413521.60 & \textbf{7413161.00} \\
17    & 8742185.00 & 8740574.00 & \textbf{8739996.50} & 8741094.00 \\
18    & 9880458.40 & 9875796.80 & \textbf{9875025.60} & 9877518.00 \\
19    & 8930651.80 & \textbf{8924365.80} & \textbf{8924365.80} & 8925508.20 \\
20    & 9380007.00 & 9377330.50 & \textbf{9376535.50} & 9376771.00 \\
21    & 9814944.80 & 9814093.00 & \textbf{9813795.00} & 9814210.60 \\
22    & 10142751.40 & 10137340.00 & 10135155.80 & \textbf{10134116.20} \\
23    & 10621326.00 & 10617706.50 & 10616418.50 & \textbf{10610939.50} \\
24    & 11077982.20 & 11073976.40 & 11073201.60 & \textbf{11067852.80} \\
25    & 12404914.80 & 12397736.20 & 12397389.40 & \textbf{12396503.20} \\
26    & 13681913.00 & 13678359.50 & 13675401.00 & \textbf{13673100.00} \\
27    & 14802946.00 & 14799642.60 & 14797663.40 & \textbf{14790927.60} \\
28    & 3194939.60 & \textbf{3190687.20} & \textbf{3190687.20} & 3191008.80 \\
29    & 3470471.50 & 3468225.50 & \textbf{3467825.50} & 3467885.50 \\
30    & 3675663.00 & 3674559.40 & 3674364.60 & \textbf{3674013.40} \\
31    & 3989579.00 & 3987901.00 & \textbf{3987305.40} & 3987614.20 \\
32    & 4492250.00 & 4491343.50 & \textbf{4490783.50} & 4491223.00 \\
33    & 4880649.80 & 4880362.80 & \textbf{4880212.60} & 4880360.80 \\
34    & 5828545.20 & 5828040.40 & 5827417.00 & \textbf{5822312.80} \\
35    & 7115807.50 & 7114859.50 & \textbf{7114010.50} & 7114766.50 \\
36    & 8110102.40 & 8109727.20 & \textbf{8109088.80} & 8109140.20 \\
37    & 4837334.80 & 4835299.20 & \textbf{4832416.20} & 4832610.20 \\
38    & 5086428.00 & 5084549.00 & 5083203.50 & \textbf{5083190.00} \\
39    & 5295171.40 & 5294310.00 & \textbf{5293782.40} & 5294049.20 \\
40    & 5824516.80 & 5821000.20 & \textbf{5818315.80} & 5819894.00 \\
41    & 6339034.00 & 6334971.50 & \textbf{6332707.50} & 6333348.50 \\
42    & 6710371.40 & 6707865.80 & \textbf{6707023.20} & 6707712.80 \\
43    & 7442889.60 & 7439899.40 & \textbf{7439000.20} & 7439542.80 \\
44    & 8718339.00 & 8717943.00 & \textbf{8716292.50} & 8716846.50 \\
45    & 9785732.00 & 9783657.60 & 9782921.00 & \textbf{9782390.60} \\
46    & 8952738.60 & 8947230.60 & \textbf{8945587.60} & 8947390.40 \\
47    & 9389338.50 & 9386929.00 & 9386828.50 & \textbf{9386696.00} \\
48    & 9841728.00 & 9841518.80 & 9841518.80 & \textbf{9841486.20} \\
49    & 10120022.20 & 10113806.60 & 10113232.60 & \textbf{10111443.20} \\
50    & 10617420.50 & 10612812.50 & \textbf{10610442.50} & 10612478.50 \\
51    & 11055554.00 & 11053156.40 & 11052142.60 & \textbf{11052075.20} \\
52    & 12410460.80 & 12398020.20 & \textbf{12396783.80} & 12397352.60 \\
53    & 13644827.50 & 13636869.50 & \textbf{13634787.00} & 13637640.50 \\
54    & 14678743.20 & 14672016.20 & \textbf{14671334.20} & 14672045.40 \\
55    & 3276712.20 & 3272956.00 & \textbf{3272178.20} & 3272501.20 \\
56    & 3521771.50 & 3520417.00 & \textbf{3519533.00} & 3519718.50 \\
57    & 3734584.00 & 3733626.60 & \textbf{3732920.00} & 3733139.00 \\
58    & 4030910.60 & 4028775.20 & \textbf{4028192.00} & 4028278.80 \\
59    & 4505687.50 & 4504917.00 & \textbf{4504624.50} & 4504865.00 \\
60    & 4890417.00 & 4889997.80 & 4889780.40 & \textbf{4889611.60} \\
61    & 5719050.40 & 5717007.00 & 5716875.00 & \textbf{5716329.60} \\
62    & 7002678.50 & 7001739.50 & 7001623.00 & \textbf{7001599.5}0 \\
63    & 8089596.80 & 8089375.80 & 8089077.80 & \textbf{8088910.20} \\
64    & 4858676.40 & 4854893.80 & 4853546.80 & \textbf{4852214.40} \\
65    & 5085849.50 & 5083797.50 & 5083342.00 & \textbf{5082806.50} \\
66    & 5275865.60 & 5275031.80 & 5274285.80 & \textbf{5274003.00} \\
67    & 5854389.60 & 5852421.80 & 5850146.40 & \textbf{5849751.40} \\
68    & 6327424.50 & 6324506.00 & 6324048.00 & \textbf{6321986.00} \\
69    & 6704104.20 & 6701988.40 & 6701984.40 & \textbf{6700566.20} \\
70    & 7414142.40 & 7412435.60 & 7412435.60 & \textbf{7412131.20} \\
71    & 8724241.50 & 8722788.50 & \textbf{8722052.50} & 8722781.50 \\
72    & 9846869.80 & 9845927.60 & 9845331.80 & \textbf{9842907.80} \\
73    & 8917010.20 & 8910955.40 & 8910448.00 & \textbf{8909969.40} \\
74    & 9398291.50 & 9396151.50 & 9395864.50 & \textbf{9395172.00} \\
75    & 9838522.00 & 9837378.40 & 9837137.00 & \textbf{9836592.80} \\
76    & 10116626.20 & 10112241.20 & 10110801.20 & \textbf{10105325.20} \\
77    & 10619852.00 & 10617052.00 & 10616872.50 & \textbf{10614811.50} \\
78    & 11089695.80 & 11088779.40 & 11088183.20 & \textbf{11086320.60} \\
79    & 12375946.80 & 12372733.20 & 12369941.00 & \textbf{12369455.40} \\
80    & 13634882.50 & 13627465.50 & 13625726.50 & \textbf{13622603.50} \\
81    & 14777131.20 & 14771777.60 & 14768766.80 & \textbf{14768413.20} \\
82    & 3214135.80 & 3210641.40 & 3210264.40 & \textbf{3209582.60} \\
83    & 3477940.00 & 3475430.50 & 3475217.50 & \textbf{3474652.50} \\
84    & 3675077.60 & 3673623.80 & 3673088.80 & \textbf{3672828.20} \\
85    & 3982416.40 & 3980182.60 & 3979697.40 & \textbf{3979655.60} \\
86    & 4505007.50 & 4503753.50 & \textbf{4503380.00} & 4503516.00 \\
87    & 4893901.60 & 4893628.60 & \textbf{4892244.60} & 4893691.60 \\
88    & 5721915.80 & 5720885.00 & \textbf{5720609.80} & 5720862.00 \\
89    & 7067795.00 & 7067129.50 & \textbf{7067017.50} & 7067166.00 \\
90    & 8095371.80 & 8095077.60 & \textbf{8094495.60} & 8094637.40 \\
91    & 4832667.00 & 4830289.40 & 4828987.40 & \textbf{4827664.60} \\
92    & 5084756.50 & 5082263.00 & 5082076.50 & \textbf{5081173.00} \\
93    & 5300186.00 & 5298852.40 & 5298269.40 & \textbf{5297930.20} \\
94    & 5815482.80 & 5811483.40 & \textbf{5809490.60} & 5809803.80 \\
95    & 6319621.50 & 6315565.50 & \textbf{6314004.00} & 6314236.00 \\
96    & 6690277.40 & 6687983.00 & \textbf{6685626.20} & 6686929.00 \\
97    & 7413442.20 & 7412557.80 & \textbf{7411258.80} & 7411695.40 \\
98    & 8697765.00 & 8696213.50 & 8695934.50 & \textbf{8694725.00} \\
99    & 9768606.20 & 9767458.00 & 9766777.00 & \textbf{9766007.20} \\
100   & 8930673.60 & 8925828.80 & \textbf{8922872.20} & 8924646.00 \\
101   & 9383498.00 & 9381106.00 & 9380728.00 & \textbf{9378242.50} \\
102   & 9828182.60 & 9826930.80 & 9826716.40 & \textbf{9826028.40} \\
103   & 10121253.40 & 10117248.80 & \textbf{10113658.20} & 10114240.40 \\
104   & 10602131.00 & 10600187.00 & \textbf{10597353.50} & 10599748.50 \\
105   & 11067832.00 & 11064491.40 & \textbf{11063036.20} & 11064491.40 \\
106   & 12396198.40 & 12394123.40 & 12392742.00 & \textbf{12391271.20} \\
107   & 13606139.00 & 13598050.50 & \textbf{13595986.50} & 13600035.50 \\
108   & 14660716.80 & 14659042.80 & 14656880.00 & \textbf{14652956.00} \\
\hline
\end{longtable}
}

These results show that, the parallel versions achieved better quality of solutions compared to the sequential results of the VND implementation, in general. Moreover, the hybrid OpenMP - OpenACC version performed slightly better than the other two parallelization schemes. These results also demonstrate that, the computational improvement is a viable goal for researchers even with the addition of certain programming directives into compute intensive loops and not based on a complete rewriting of the code.

%-------------------------------------------------------------------------------------------
%-------------------------------------------------------------------------------------------

\section{Conclusions and Future work}
\label{conclusions}

\vspace{-0.2cm}

In this paper a new hybrid CPU-GPU GVNS metaheuristic algorithm based on OpenMP and OpenACC programming languages was proposed. The algorithm was applied on a recent set of 108 benchmark instances of a hard inventory optimization problem and produced better quality solutions not only from the corresponding sequential version but also from other two parallel implementations. Moreover, it would be interesting to use other programming languages either for CPU parallel programming (i.e., MPI) or for GPU parallel programming (i.e., CUDA, OpenCL), in order to assess their benefits.

\vspace{-0.3cm}

%-------------------------------------------------------------------------------------------
%-------------------------------------------------------------------------------------------

\section*{Acknowledgement}
\label{Acknowledgement}

\vspace{-0.3cm}

We gratefully acknowledge the support of NVIDIA Corporation with the donation of the GTX 980 GPU used for this research.

\vspace{-0.3cm}

\bibliographystyle{endm}
\bibliography{VNS2016}

\end{document}